\def\year{2019}\relax
\documentclass[letterpaper]{article} 
\usepackage{aaai19}  
\usepackage{times}  
\usepackage{helvet}  
\usepackage{courier}  
\usepackage{url}  
\usepackage{graphicx}  

\usepackage{amsfonts} 
\usepackage{nicefrac} 
\usepackage{microtype} 

\usepackage[tbtags]{amsmath}
\usepackage{multirow}
\usepackage{listings}
\usepackage{color}

\begin{document}
%
\title{Knowledge-incorporating ESIM models for Response Selection in Retrieval-based Dialog Systems}
\author{
  Jatin Ganhotra \\
  IBM Research \\
  {\tt jatinganhotra@us.ibm.com}
  \\\And
  Siva Sankalp Patel \\
  IBM Research \\
  {\tt siva.sankalp.patel@ibm.com}
  \\\And
  Kshitij Fadnis \\
  IBM Research \\
  {\tt kpfadnis@us.ibm.com}
}

\maketitle

\begin{abstract}
Goal-oriented dialog systems, which can be trained end-to-end without manually encoding domain-specific features, show tremendous promise in the customer support use-case e.g. flight booking, hotel reservation, technical support, student advising etc. These dialog systems must learn to interact with external domain knowledge to achieve the desired goal e.g. recommending courses to a student, booking a table at a restaurant etc. This paper presents extended Enhanced Sequential Inference Model (ESIM) models: a) K-ESIM (Knowledge-ESIM), which incorporates the external domain knowledge and b) T-ESIM (Targeted-ESIM), which leverages information from similar conversations to improve the prediction accuracy. Our proposed models and the baseline ESIM model are evaluated on the Ubuntu and Advising datasets in the \textit{Sentence Selection} track of the latest Dialog System Technology Challenge (DSTC7), where the goal is to find the correct next utterance, given a partial conversation, from a set of candidates. Our preliminary results suggest that incorporating external knowledge sources and leveraging information from similar dialogs leads to performance improvements for predicting the next utterance.
\end{abstract}

\section{Introduction}
\label{Introduction}
The \textit{Dialog State Tracking Challenge} (DSTC) was initially started to serve as a testbed for dialog state tracking task and was rebranded as \textit{Dialog System Technology Challenges} in 2016 to provide a benchmark for evaluating dialog systems. The latest DSTC challenge, DSTC7\footnote{\url{http://workshop.colips.org/dstc7/index.html}}, is divided into three different tracks. Our work addresses the \textit{Sentence Selection} track, where the objective is to push the utterance classification towards real world problems.

Retrieval-based dialog systems, in comparison to generation-based dialog systems, have the benefit of informative and fluent responses, as the proper response for the current conversation is selected from a set of candidates responses. The main goal for a dialog system in the Sentence Selection track is that it should:
\begin{itemize}
    \item select the correct next utterance(s) from a set of candidates, given a partial conversation.
    \item learn to select none of the candidates if none of them align with the given partial conversation.
    \item learn to incorporate external knowledge sources.
\end{itemize}

Goal-oriented dialog systems usually rely on external knowledge sources, such as restaurant names and details for restaurant table booking task, course information at a university for student advising task and flight details for flight reservation task. The external knowledge could be provided in a structured format (a knowledge base (KB)) or unstructured format (man pages for Linux commands). It is essential for a dialog system to incorporate the external knowledge source for task completion and achieving the desired goal. \citeauthor{bordes2016learning} \shortcite{bordes2016learning} introduced bAbI dialog tasks, as a simulation for restaurant table booking task but interaction with the domain KB was circumvented as restaurant details relevant to a given dialog were provided as part of the dialog history. \citeauthor{eric2017key} \shortcite{eric2017key} proposed Key-Value retrieval networks, which performs attention over the entries of a KB to extract relevant information. \citeauthor{lowe2015incorporating} \shortcite{lowe2015incorporating} explored incorporating unstructured external textual information (man pages) for the next utterance classification task on Ubuntu dialog corpus. In this work, we extend the Enhanced Sequential Inference Model (ESIM) \cite{chen2017enhanced} and propose two end-to-end models for next utterance selection:

\begin{itemize}
    \item K-ESIM (Knowledge-based ESIM) incorporates the additional unstructured external textual information.
    \item T-ESIM (Targeted ESIM) leverages information from similar dialogs seen during training.
\end{itemize}

The paper is structured as follows. In the next section, we briefly describe the problem and the two datasets used for evaluating our proposed models. Then, we present related work and introduce the baseline ESIM model and our proposed models: K-ESIM and T-ESIM. In the Experiments and Results section, we present our evaluation results across all models on the two datasets. Finally, we draw conclusion from our work and indicate directions of future work.

\section{Problem Statement and Datasets}
\label{Problem-Statement-and-datasets}
The DSTC7 challenge for \textit{Sentence Selection} track consists of two datasets: a) Ubuntu dataset and b) Advising dataset. Both datasets share the common goal to predict the correct next utterance from a set of potential next utterance candidates, given a partial conversation. In addition to selecting the correct next utterance, the dialog system is also evaluated on its ability to identify that none of provided candidates is a good next utterance for the given partial conversation. The 5 subtasks and the 2 datasets are described below:
\begin{itemize}
    \item \textbf{Subtask 1}: The Subtask 1 serves as the baseline task for both datasets. The goal is to select the next utterance for the partial conversation from the given candidate set, which contains 1 correct and 99 incorrect next utterances.
    
    \item \textbf{Subtask 2}: Subtask 2 increases the task complexity by testing the dialog system to select the next utterance for the partial conversation from a large global pool of next utterance candidates. Subtask 2 is evaluated only on the Ubuntu dataset, where the total number of candidates in the global pool is 120000.
    
    \item \textbf{Subtask 3}: The goal of Subtask 3 is to evaluate the dialog system to select all correct next utterances from the given set of candidates. Subtask 3 is evaluated only on the Advising dataset, where the given candidate set can contain 1 to 5 correct next utterances (original correct utterance and paraphrases) and 95 to 99 incorrect next utterances.
    
    \item \textbf{Subtask 4}: Subtask 4 extends Subtask 1 and serves as a benchmark to check if the dialog system can learn to identify when correct next utterance for the partial conversation is not available in the given candidate set. For such cases, the dialog system must respond with '\textit{None}' as next utterance. Subtask 4 is evaluated on both datasets.
    
    \item \textbf{Subtask 5}: In Subtask 5, the dialog system is evaluated on its ability to incorporate additional external knowledge provided and is evaluated on both datasets. The expectation for Subtask 5 is that the dialog system must perform better after incorporating external knowledge.
\end{itemize}

\subsection{Ubuntu dialog corpus}
\label{Ubuntu-dialog-corpus}
The Ubuntu dialog corpus provided as part of the challenge is a new version of disentangled two-party conversations from Ubuntu IRC logs \cite{kummerfeld2018analyzing}. The purpose is to solve an Ubuntu user's posted problem, i.e. select the correct next utterance based on the conversation so far between the two users. The training data contains over 100k complete conversations, and the test data contains 1000 partial conversations, where each dialog has a minimum of 3 turns. In addition to the dialog corpus, additional knowledge is also provided in the form of linux manual pages.

\subsection{Advising dataset}
\label{Advising-dataset}
The Advising data contains conversations between a student and advisor, where the goal of the advisor is to guide the student to pick courses that a) align with the student's curriculum and b) match the student's personal preferences about time (when classes are held e.g. morning, afternoon etc.), course difficulty (easy, hard etc.), career path etc. The dataset collected is play-acted where two students act as the two roles using provided personas. The data also includes paraphrases of the sentences and of the target responses by the advisor. In addition to the dataset, an additional knowledge base (KB) is provided which contains information about various courses and possible personal preferences for the students. The training data contains 100,000 partial dialogs from the original 500 dialogs. The test data consists of 500 partial dialogs, where a set of 100 candidates are provided which includes 1-5 correct next utterances.

Additional details for the Ubuntu and Advising datasets, along with the external knowledge sources for both datasets are provided in the \textit{Sentence Selection} track description document\footnote{\url{https://ibm.github.io/dstc7-noesis/public/data_description.html}}.

\begin{figure*}[h]
\centering
\includegraphics[width=0.98\textwidth]{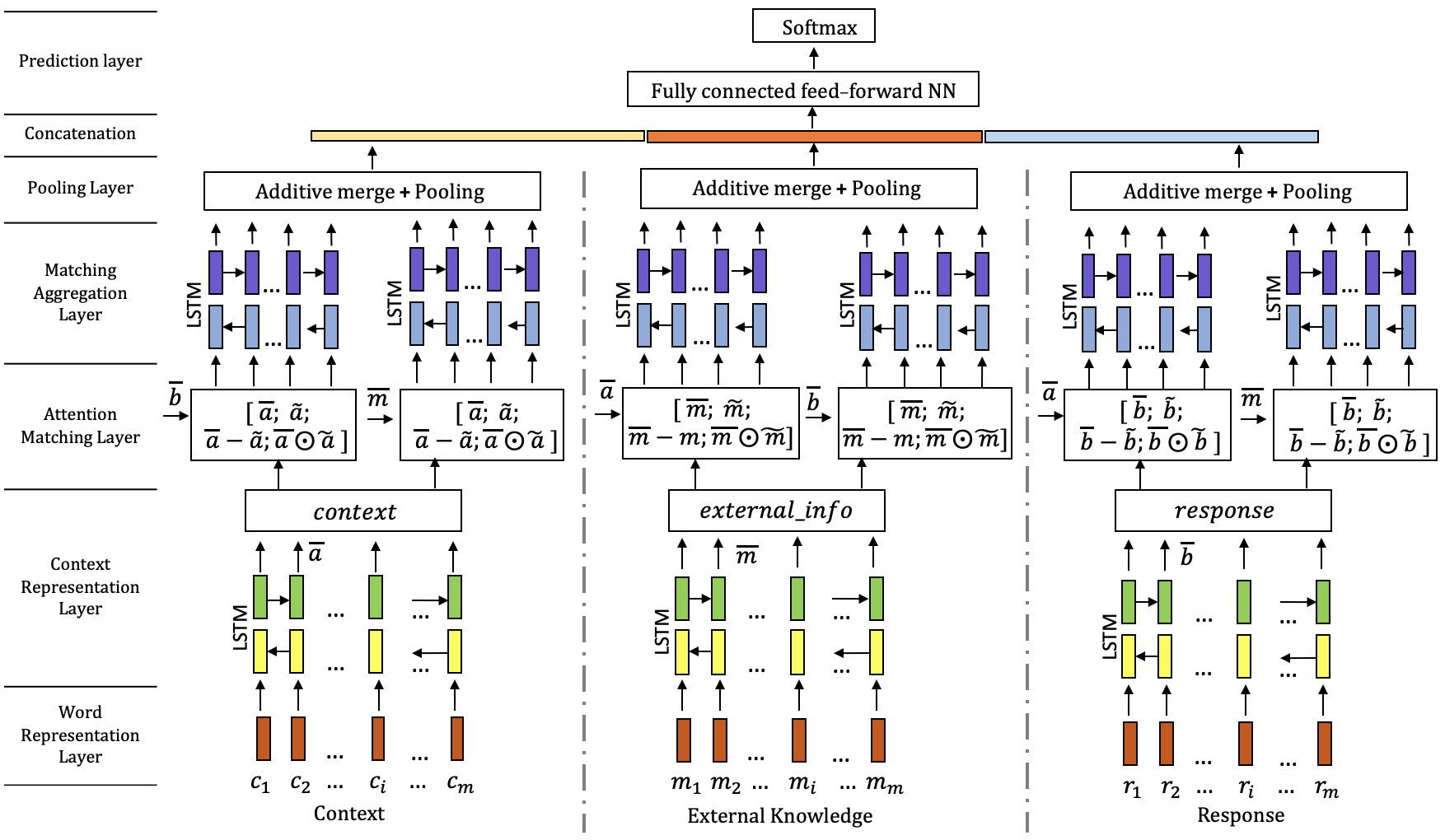}
\caption{\textbf{K-ESIM}: A high-level overview of K-ESIM model, which incorporates external knowledge.}
\label{fig_k_esim}
\end{figure*}

\section{Related Work}
\label{related-work}
End-to-end dialog systems, based on neural networks, have shown the promise of learning directly from human-to-human dialog interactions. They have performed well in non goal-oriented chit-chat settings (
\citeauthor{vinyals2015neural} \shortcite{vinyals2015neural}; \citeauthor{sordoni2015neural} \shortcite{sordoni2015neural}; \citeauthor{serban2016building} \shortcite{serban2016building};
as well as goal-oriented settings (\citeauthor{le2016lstm} \shortcite{le2016lstm}, \citeauthor{ghazvininejad2017knowledge} \shortcite{ghazvininejad2017knowledge},  \citeauthor{bordes2016learning} \shortcite{bordes2016learning}, \citeauthor{seo2016query} \shortcite{seo2016query}). There are two active research directions for building goal-oriented dialog systems: \textit{generative}, where the system generates the next utterance in the conversation word-by-word, and \textit{retrieval-based}, where the system has to pick the next utterance from a list of potential responses. Response selection has been actively explored by the research community in the last few years (\citeauthor{dong2018enhance} \shortcite{dong2018enhance}, \citeauthor{wu2016sequential} \shortcite{wu2016sequential}, \citeauthor{bartl2017retrieval} \shortcite{bartl2017retrieval}).

Ubuntu dialog corpus introduced by \citeauthor{lowe2015ubuntu} \shortcite{lowe2015ubuntu} consists of conversations from the Ubuntu IRC channel where users ask and answer technical questions about Ubuntu. The diversity of conversations and large quantity of dialogs available in the corpus presents a unique challenge for goal-oriented dialog systems. \citeauthor{lowe2015ubuntu} \shortcite{lowe2015ubuntu} proposed Dual-encoder architecture which uses Long Short-Term Memory (LSTM) \cite{hochreiter1997long} to embed both context and response into vectors and response selection is based on the similarity of embedded vectors. \citeauthor{kadlec2015improved} \shortcite{kadlec2015improved} built an ensemble of convolution neural network (CNN) \citeauthor{krizhevsky2012imagenet} \shortcite{krizhevsky2012imagenet} and Bi-directional LSTM. 
\citeauthor{dong2018enhance} \shortcite{dong2018enhance} integrated character embeddings \cite{dos2015boosting} into Enhanced LSTM method (ESIM) \cite{chen2017enhanced} and achieve significant performance improvement.

There have been several studies on incorporating unstructured external information into dialog models. \citeauthor{ghazvininejad2017knowledge} \shortcite{ghazvininejad2017knowledge} proposed a knowledge-grounded neural conversation model by improving the seq2seq approach to produce more contentful responses. The model was trained using Twitter Dialog Dataset \cite{li2016diversity} as dialog context and foursquare data as the external information. \citeauthor{young2017augmenting} \shortcite{young2017augmenting} incorporated structured knowledge from knowledge graphs and achieved an improved dialog system over the Twitter Dialog Dataset.  \citeauthor{lowe2015incorporating} \shortcite{lowe2015incorporating} used the Linux manual pages as the external knowledge information for improving the next utterance prediction task and showed reasonable accuracy gains.


\section{Baseline model: ESIM}
\label{baseline-model-esim}

We use the ESIM model proposed by \citeauthor{chen2017enhanced} \shortcite{chen2017enhanced} as the baseline model. The implementation details for the baseline model are provided in Appendix. As mentioned in the 'Problem Statement' section, the task is to select the next response given the dialog history (context). The multi-turn dialog history is concatenated together to form the context of length $m$, represented as $C = (c_1, c_2, ..., c_i, ..., c_m)$, where $c_i$ is the $i$th word in context. Given a response $R$ with length $n$ as $R = (r_1, r_2, ..., r_j, ..., r_n)$ where $r_j$ is the $j$th word in response, the next response is 
selected using the conditional probability $P(y=1|C,R)$, which shows the confidence of selecting the response $R$ given context $C$. 

\subsection{Word Representation Layer:}
We represent each word in the context and the response with a $d$-dimension vector. Following \citeauthor{dong2018enhance} \shortcite{dong2018enhance}, the word representation is generated by concatenating the word embedding and the character-composed embedding for the word.  The word embedding for each word is generated by concatenating the Glove word embeddding \cite{pennington2014glove} and the word2vec embedding \cite{mikolov2013efficient} for the word. The word2vec vectors are generated by training on the training data, where each sentence is treated as a document. The character-composed embedding, introduced by \cite{dong2018enhance} is generated by concatenating the final state vector of the forward and backward direction of bi-directional LSTM (BiLSTM).

\subsection{Context representation layer:}
The context representation layer utilizes BiLSTM to capture word representation and it's local sequence context. The hidden states at each time step for both directions are concatenated to form a new local context-aware word representation, denoted by $\bar{a}$ and $\bar{b}$ for context and response 
below.

\begin{equation}
\label{eqContextRepresentationLayer1}
\bar{a_i} = BiLSTM(\bar{a}_{i-1}, w_i), 1 \leq i \leq m
\end{equation}
\begin{equation}
\label{eqContextRepresentationLayer2}
\bar{b_j} = BiLSTM(\bar{b}_{j-1}, w_j), 1 \leq j \leq n
\end{equation}

\subsubsection{Attention matching layer:}
As in ESIM model, the co-attention matrix where $E \in \mathbb{R}^{m*n}$ where $E_{ij} =\bar{a}_i^T\bar{b}_j$ computes the similarity between context and response. The attended response vector computed via equation \ref{eqAttentionMatchingLayer1} represents the most relevant response word for each word in context. Similarly, the attented context vector is computed via equation \ref{eqAttentionMatchingLayer2} and represents the most relevant context word for each word in response.
\begin{equation}
\label{eqAttentionMatchingLayer1}
\tilde{a_i} = \sum_{j=1}^n \frac{exp(E_{ij})}{\sum_{k=1}^n exp(E_{ik})}\bar{b}_j, 1 \leq i \leq m
\end{equation}
\begin{equation}
\label{eqAttentionMatchingLayer2}
\tilde{b_j} = \sum_{i=1}^m \frac{exp(E_{ij})}{\sum_{k=1}^m exp(E_{kj})}\bar{a}_i, 1 \leq j \leq n
\end{equation}

Using the attended context and response vectors from equations \ref{eqAttentionMatchingLayer1} and \ref{eqAttentionMatchingLayer2}, vector difference and element-wise product is computed
to further elicit interaction information between context and response. The difference and element-wise product vectors are concatenated with the original vectors to generate $m_a^i$ and $m_b^i$ as shown below. 
\begin{gather}
\label{eq:5}
m_a^i = [\bar{a_i},\tilde{a_i};\bar{a_i}-\tilde{a_i}; \bar{a_i}\odot\tilde{a_i}], 1 \leq i \leq m \\
\label{eq:6}
m_b^i = [\bar{b_i},\tilde{b_i};\bar{b_i}-\tilde{b_i}; \bar{b_i}\odot\tilde{b_i}], 1 \leq j \leq n
\end{gather}

\subsection{Matching aggregation layer:}
In this layer, another BiLSTM is used to aggregate response-aware context representations and context-aware response representations \cite{chen2017enhanced}. The matching aggregation layer learns to compose local inference information sequentially using the BiLSTM, as shown below.
\begin{equation}
\label{eq:7}
v_i^a = BiLSTM(v_{i-1}^a,m_i^a), 1\leq i \leq m,
\end{equation}
\begin{equation}
\label{eq:8}
v_j^b = BiLSTM(v_{j-1}^b,m_j^b), 1\leq j \leq n.
\end{equation}
\subsection{Pooling layer:}
We use max pooling via combining max pooling and final state vectors (concatenation of both forward and backward one) ($v_{last}^a ; v_{last}^b$) to form the final fixed vector \cite{dong2018enhance}, which is calculated as follows:
\begin{equation}
\label{eq:9}
v_{max}^a = \max_{i=1}^m v_i^a
\end{equation}
\begin{equation}
\label{eq:10}
v_{max}^b = \max_{j=1}^n v_j^b
\end{equation}
\begin{equation}
\label{eq:11}
v = [v_{max}^a;v_{max}^b;v_{last}^a;v_{last}^b]
\end{equation}
$v$ from equation \ref{eq:11}, is fed into the final prediction layer (a fully-connected feed-forward neural network).

\section{Proposed model: K-ESIM}
\label{proposed-model-k-esim}
We use the baseline ESIM model described above and update the model architecture for incorporating external knowledge e.g. command description from man pages for Ubuntu dataset. We refer to the new model as K-ESIM, (Knowledge incorporating ESIM model). A high-level overview of K-ESIM is shown in Figure \ref{fig_k_esim}. We introduce the following changes to the baseline ESIM model:

\subsection{Word Representation and Context Representation layers:}
The external command information is passed through the same \textit{Word Representation} and \textit{Context Representation} layers i.e. these layers share weights for encoding a) the dialog context, b) the candidate response and c) the external knowledge. We also experimented with using separate weights for external knowledge (weight untying) i.e. a different BiLSTM was used to encode external information but did not observe increase in performance.

\subsection{Attention Matching layer:}
The Attention Matching layer is updated to incorporate the additional external information, such as man pages for Ubuntu dataset and course information for Advising dataset. In addition to computing co-attention between the dialog history (context) and the candidate response, we also compute attention between a) the dialog context and the external information and b) candidate response and the external information. Using the attention scores for similarity, we compute the following:
\begin{itemize}
    \item attended context vectors (based on candidate response and the external information) 
    \item attended candidate response vectors (based on context and the external information) 
    \item attended external information vectors (based on context and candidate response)
\end{itemize}
These attended vectors are then used to compute vector difference and element-wise product, to enrich the interaction information between each pair from the set \{dialog context, candidate response, external information\}, similar to equations \ref{eq:5} and \ref{eq:6} as shown in Figure \ref{fig_k_esim}.

\subsection{Matching Aggregation and Pooling layers:}
The Matching Aggregation layer aggregates the encoded information from \textit{Attention Matching} layer. We use the same BiLSTM for all pairs mentioned above (weight-tying). Since we have 2 representations for each variable \{dialog context, candidate response and external information\}, we perform an additive merge in the \textit{Pooling layer} to get a final representation for each variable, which gets concatenated and used as input to the final \textit{Prediction layer}.

\begin{table*}[htbp!]
\begin{center}
\begin{tabular}{lcccc|cccc}
    \hline
    \multirow{2}{*}{\textbf{Model}} & \multicolumn{4}{c}{\textbf{Validation}} & \multicolumn{4}{c}{\textbf{Test}}\\
     & R@1 & R@10 & R@50 & MRR & R@1 & R@10 & R@50 & MRR  \\ \hline
    \multicolumn{9}{c}{\textbf{subtask-1}}\\\hline
    ESIM & 43.76 & 71.70 & 95.84 & 53.24 &  50.1 & 78.3 & 95.4 & 59.34 \\ \hline
    T-ESIM & 52.62 & 76.46 & 96.08 & 60.57 & 61.9 & 82.2 & 96.6 & 69.09\\ \hline
    T-ESIM-CR & 54.46 & 79.26 & 97.92 & 62.73 & 63.4 & 84.2 & 98.5 & 70.69\\ \hline
    T-ESIM-Sampled & 53.16 & 78.5 & 96.46 & 61.54 & 62.8 & 83.4 & 96.6 & 69.7\\ \hline
    T-ESIM-Sampled-CR & 55.46 & 81.98 & 98.2 & 64.12 & 64.3 & 84.7 & 97.3 & 71.25\\ \hline
    \multicolumn{9}{c}{\textbf{subtask-2}}\\\hline
    ESIM & 11.12 & 31.5 & 58.6 & 18.59 &  12.8 & 28.5 & 36.5 & 18.43 \\ \hline
    T-ESIM & 18.72 & 35.9 & 61.26 & 25.13 & 21.6 & 36.0 & 44.1 & 26.68 \\ \hline
    \multicolumn{9}{c}{\textbf{subtask-4}}\\\hline
    ESIM & 40.16 & 76.18 & 96.36 & 53.43 &  43.5 & 82.1 & 96.2 & 57.96 \\ \hline
    T-ESIM & 47.76 & 77.22 & 96.4 & 58.43 & 52.5 & 82.3 & 97.1 & 63.6 \\ \hline
    \multicolumn{9}{c}{\textbf{subtask-5}}\\\hline
    K-ESIM & 44.82 & 72.74 & 96.4 & 54.52 &  50.1 & 78.3 & 96.3 & 60.2 \\ \hline
    TK-ESIM & 53.10 & 75.88 & 96.26 & 60.88 & 60.9 & 80.2 & 96.6 & 67.93 \\ \hline
    TK-ESIM-CR & 54.84 & 79.26 & 97.96 & 62.98 & 62.3 & 83.4 & 97.8 & 69.56\\ \hline
    
\end{tabular}
\end{center}
\caption{Performance of models on the Ubuntu validation and test datasets. R@k refers to Recall at position k in 100 candidates, denoted as R@1, R@10 and R@50. MRR refers to the Mean Reciprocal Rank.}
\label{tab:baseline-results-ubuntu}
\end{table*}


\begin{table*}[h]
\begin{center}
\begin{tabular}{lcccc|cccc}
    \hline
    \multirow{2}{*}{\textbf{Model}} & \multicolumn{4}{c}{\textbf{Validation}} & \multicolumn{4}{c}{\textbf{Test}}\\
     & R@1 & R@10 & R@50 & MRR & R@1 & R@10 & R@50 & MRR  \\ \hline
    ESIM (subtask-1) & 17.2 & 47.6 & 88.8 & 27.5 &  14.8 & 46.2 & 86.6 & 25.43 \\ \hline
    ESIM (subtask-3) & 10.2 & 47.6 & 87.6 & 22.08 &  18.6 & 60.2 & 92.6 & 31.62 \\ \hline
    ESIM (subtask-4) & 22.2 & 57.2 & 91.8 & 33.89 &  17.0 & 72.8 & 91.2 & 30.14 \\ \hline
    K-ESIM (subtask-5) & 16.4 & 50.4 & 85.6 & 27.45 &  11.6 & 49.2 & 88.2 & 23.02 \\ \hline
\end{tabular}
\end{center}
\caption{Performance of models on the Advising validation and test datasets. R@k refers to Recall at position k in 100 candidates, denoted as R@1, R@10 and R@50. MRR refers to the Mean Reciprocal Rank.}
\label{tab:baseline-results-advising}
\end{table*}

\subsection{Extracting relevant external information:}
\textbf{Ubuntu dataset:} The external data is provided as man pages information for 1200 Linux commands.
We map the command description for all commands into TF-IDF vectors. We use two hashtables: entity hashtable and relation hashtable for knowledge extraction \cite{lowe2015incorporating}. The command names are used as the keys to the entity hashtable, and their descriptions as the corresponding values. The words following the command name in the \textit{Name} section\footnote{The Name section has a one-sentence summary of the command} are used as keys for the relation hashtable. We also filter the keys of the relation hashtable by removing common English words. The values to the relation hashtable contains the commands from which the keys of the relation hashtable were extracted. Thus, the relation hashtable contains pointers to the keys in the entity hashtable. 

We compare all the tokens in the context with keys in entity hashtable to find any direct matches for command names. We also check for partial matches if the token length is greater than 8 as sometimes the full command name may not be used in the context. If we don't get any matches in the entity hashtable, we look for token matches in the context with keys in the relation hashtable. 
These matches eventually lead us to the list of command names relevant to the given dialog. 
The shortlisted commands are ranked based on the TF-IDF match score for their corresponding description with the given dialog context. 
We further split the top-k\footnote{We used k=5 in our experiments.} command descriptions into individual sentences. These sentences are again ranked in descending order of their cosine similarity with respect to the context. Finally, first 200 words were selected from these ranked sentences\footnote{We select the first 200 words as we use num\_units=200 for the BiLSTM cells in our model.}.

\textbf{Advising dataset:} The external data is provided in a KB which contains information such as `Course Name', `Area', `Credits', `Workload', `Class Size', etc. 
We construct a natural language sentence representation for the course information provided in the KB using a set of predefined rules. 
Each suggested course is mapped to its corresponding sentence representation. An example natural language sentence representation for a suggested course is: `` \texttt{EECS281} is \texttt{Data Structures and Algorithms}, has \texttt{moderate} workload,  \texttt{large} class size, \texttt{4} credits, has \texttt{a discussion}, the classes are \texttt{on Thursday, Tuesday afternoon}''. This representation is used as external knowledge input to our proposed model K-ESIM for the given dialog.

\section{Proposed model: T-ESIM}
\label{proposed-model-t-esim}
In a dialog corpus, similar conversations can appear many times. For example, in a customer care scenario, a common problem might arise for multiple users and many similar dialogs would be present in the corresponding dialog corpus. \citeauthor{pandey2018exemplar} \shortcite{pandey2018exemplar} proposed Exemplar Encoder-Decoder (EED) architecture that makes use of similar conversations for the generation-based dialog system and achieved better results than models such as HRED \cite{serban2016building} and VHRED \cite{serban2017hierarchical}. We adpot a similar approach and propose a new training strategy to incorporate the information from similar dialogs present in the available training data for the next utterance selection task. We refer to the new training strategy as T-ESIM (Targeted ESIM), where additional information in terms of probable target responses is added to the contextual information. 


For our T-ESIM implementation, we use text-based similarly technique to identify relevant dialogs, similar to K-ESIM. Each dialog in the training data is split at multiple points to create a larger pool of dialogs, which are called sub-dialogs. The sub-dialogs are then converted to TF-IDF vector representations. The current dialog is matched against these sub-dialogs (excluding its children) to identify similar sub-dialogs to select top-k similar sub-dialogs\footnote{We use k=3 during training and k=1 during evaluation}. The corresponding response(s) for the top-k similar sub-dialogs are concatenated to the current dialog context as a new turn in the partial conversation\footnote{The training data is increased by a factor of k}. The core motivation here is that the model can learn to use responses for similar dialogs present in the training data, to get improved performance on the next utterance selection task. We also explore additional training strategies: T-ESIM-Sampled and T-ESIM-CR, which are described below. We evaluate these strategies on the Ubuntu dialog corpus, as shown in Table \ref{tab:baseline-results-ubuntu}.

\textbf{T-ESIM-Sampled}: When the candidate set for each dialog contains 100 utterances, 1 candidate is the correct utterance and the remaining 99 candidates are incorrect responses. To speed up the training, we randomly sample 9 utterances from the 99 incorrect utterances. We refer to this training strategy as \textit{T-ESIM-Sampled}.

\textbf{T-ESIM-CR}: The Ubuntu and Advising corpus are constructed from real-world human-to-human conversations. This makes them unique, as different people can answer the same question in different ways. The different answers could theoretically have the same information but would differ in terms of natural language. Therefore, during evaluation, we employ the Candidate Reduction (CR) trick to use the presence of unique responses in the dataset. For evaluation on the validation set, we reduce the total number of candidates in the candidate set by removing the candidates which are present as correct responses in the training data and similarly, for test data, we remove the candidates which are present in the training and validation data.

\section{Experiments and Results}
\label{Experiments and Results}
Our results for the baseline model ESIM and our proposed models: K-ESIM and T-ESIM for the Ubuntu dataset are given in Table \ref{tab:baseline-results-ubuntu} and for the Advising dataset are given in Table \ref{tab:baseline-results-advising}. The models are evaluated on two metrics - Recall@k, which refers to recall at position k in the set of the 100 candidates and MRR (mean reciprocal rank).

We observe that the baseline ESIM model achieves 50.1 R@1 on the Ubuntu test set and 14.8 R@1 on the Advising test set for subtask 1. For Advising dataset subtask 5, we observe that the K-ESIM model performance is slightly below the baseline ESIM model. We believe that our external knowledge representation for the Advising dataset is not suited for the task. For the Ubuntu dataset, we also observe that our proposed models: K-ESIM and T-ESIM perform better than the baseline ESIM model. K-ESIM achieves 44.82 R@1 and 0.5452 MRR on Ubuntu subtask 5 validation set, compared to 43.76 R@1 and 0.5324 MRR for ESIM. K-ESIM also performs slightly better than ESIM on MRR on the test set. T-ESIM performs significantly better than the baseline ESIM model on all Ubuntu subtasks and achieves 61.9 R@1 on subtask 1. Our proposed techniques T-ESIM-Sampled and T-ESIM-CR perform well and achieve 64.3 R@1 score on the Ubuntu subtask 1. These results show that our proposed models and training strategies perform well.

For Ubuntu Subtask 2, the size of global pool of candidates is 120000. For training purposes, we reduce the candidate set by randomly sampling 99 incorrect responses from the global pool. These 99 responses, in addition to the correct response, construct our candidate set of 100 responses per dialog, similar to Subtask 1. During evaluation on the validation and test sets, we first employ the CR technique mentioned above. Then, we shortlist the number of candidates to 100, by selecting the top-100 candidates from the reduced candidate global pool using IR-based methods similar to knowledge extraction for K-ESIM and T-ESIM.

\section{Conclusion and Future Work}
In this paper, we introduced two knowledge incorporating end-to-end dialog systems for retrieval-based goal-oriented dialog, by extending the ESIM model. Evaluation based on the Ubuntu dataset show that our methods are effective to improve performance by incorporating additional external knowledge sources and leveraging information from similar dialogs. Although our proposed model K-ESIM shows improvement on the Ubuntu subtask 5, we observe a slight decrease in performance on the Advising subtask 5 as explained in the previous section.

In our future work, we plan to explore the following areas to improve our proposed K-ESIM and T-ESIM models: a) improve the knowledge representation for course information, b) investigate attention mechanisms over a KB \cite{eric2017key} and c) explore neural approaches, instead of TF-IDF, for extracting relevant external information (man pages) and identifying similar dialogs for T-ESIM. 

\section{Appendix: Model Training and Hyperparameter Details}
In Word Representation Layer, we used 300-dimensional Glove pre-trained vectors\footnote{glove.42B.300d.zip : \url{https://nlp.stanford.edu/projects/glove/}} (\cite{pennington2014glove}), 100-dimensional word2vec vectors \cite{mikolov2013efficient} and 80-dimensional character-composed embedding vectors for generating the representation of a word. For training word2vec vectors, we use the \texttt{gensim.models.Word2Vec} API with the following hyper-parameters: size=100, window=10, min\_count=1 and epochs=20. The final prediction layer is a 2-layer fully-connected feed-forward neural network with ReLu activation. We use sigmoid function and minimize binary cross-entropy loss for training and updating the model.

The baseline model was implemented in Tensorflow \cite{abadi2016tensorflow} and we used the source code released by \citeauthor{dong2018enhance} \shortcite{dong2018enhance}\footnote{source code released by \citeauthor{dong2018enhance} \shortcite{dong2018enhance}: \url{https://github.com/jdongca2003/next_utterance_selection}} for the baseline model. We generated word2vec word embeddings from scratch on the DSTC7 datasets as mentioned in Algorithm-1 from \citeauthor{dong2018enhance} \shortcite{dong2018enhance}. We used Adam \cite{kingma2014adam} with a learning rate of 0.001 and exponential decay with a decay rate of 0.96 decayed every 5000 steps. Batch size used was 128. The number of hidden units for BiLSTM in both the context representation layer and the matching aggregation layer was 200. For the prediction layers, we used 256 hidden units with ReLU activation. 

\newpage
\bibliography{aaai}
\bibliographystyle{aaai}

\end{document}